\documentclass{article}

\PassOptionsToPackage{numbers, compress}{natbib}

\usepackage[preprint]{neurips_2024}

\usepackage[utf8]{inputenc} 
\usepackage[T1]{fontenc}    
\usepackage{hyperref}       
\usepackage{url}            
\usepackage{booktabs}       
\usepackage{amsfonts}       
\usepackage{nicefrac}       
\usepackage{microtype}      
\usepackage{xcolor}         
\usepackage[numbers]{natbib}
\usepackage{amsmath}    
\usepackage{amssymb}    
\usepackage{bbm}
\usepackage{graphicx}   
\usepackage{adjustbox}
\usepackage{wrapfig}

\title{Dynamic Backtracking in GFlowNets: Enhancing Decision Steps with Reward-Dependent Adjustment Mechanisms}

\author{%
  Shuai Guo\\ 
  School of Computing and Artificial Intelligence\\
  Southwest Jiaotong University\\
  Chengdu, China 611756 \\
  \texttt{guoshuai@my.swjtu.edu.cn} \\
  \And
   Jielei Chu\thanks{Corresponding author.} \\
   School of Computing and Artificial Intelligence \\
   Southwest Jiaotong University \\
   Chengdu, China 611756 \\
   \texttt{jieleichu@swjtu.edu.cn} \\
  \And
  Lin Ma\\
  chool of Transportation and Logistics\\
  Southwest Jiaotong University \\
  Chengdu, China 611756 \\
  \texttt{malin@swjtu.edu.cn}
  \And
  Zhaoyu Li\\
  School of Computing and Artificial Intelligence \\
  Southwest Jiaotong University \\
  Chengdu, China 611756 \\
  \texttt{roylzy123@163.com}
  \And
  Tianrui Li\\
  School of Computing and Artificial Intelligence \\
  Southwest Jiaotong University \\
  Chengdu, China 611756 \\
  \texttt{trli@swjtu.edu.cn} \\
}

\begin{document}

\maketitle

\begin{abstract}
Generative Flow Networks (GFlowNets or GFNs) are probabilistic models predicated on Markov flows, and they employ specific amortization algorithms to learn stochastic policies that generate compositional substances including biomolecules, chemical materials, etc. With a strong ability to generate high-performance biochemical molecules, GFNs accelerate the discovery of scientific substances, effectively overcoming the time-consuming, labor-intensive, and costly shortcomings of conventional material discovery methods. However, previous studies rarely focus on accumulating exploratory experience by adjusting generative structures, which leads to disorientation in complex sampling spaces. Efforts to address this issue, such as \textit{LS-GFN}, are limited to local greedy searches and lack broader global adjustments. This paper introduces a novel variant of GFNs, the Dynamic Backtracking GFN (DB-GFN), which improves the adaptability of decision-making steps through a reward-based dynamic backtracking mechanism. DB-GFN allows backtracking during the network construction process according to the current state's reward value, thereby correcting disadvantageous decisions and exploring alternative pathways during the exploration process. When applied to generative tasks involving biochemical molecules and genetic material sequences, DB-GFN outperforms GFN models such as \textit{LS-GFN} and \textit{GTB}, as well as traditional reinforcement learning methods, in sample quality, sample exploration quantity, and training convergence speed. Additionally, owing to its orthogonal nature, DB-GFN shows great potential in future improvements of GFNs, and it can be integrated with other strategies to achieve higher search performance.
\end{abstract}

\section{Introduction}
In the contemporary era, humanity is faced with pressing challenges such as climate crisis, pandemics, and antibiotic resistance. There is hope that these challenges can be overcome through the creation of novel biological substances and the discovery of new materials. For instance, the discovery of new materials plays a vital role in improving the efficiency of energy production and storage. Meanwhile, lowering the cost and time associated with drug discovery could potentially mitigate the impacts of emerging diseases more effectively and rapidly. However, the large costs involved in traditional material chemistry and biochemistry, along with the need for continuous trial and error in the search for high-performance substances, make the discovery process prohibitively expensive. Consequently, researchers in fields such as materials science and biochemistry are turning to use machine learning, given its ability to significantly accelerate the process of scientific discovery more efficiently and cost-effectively.

Generative Flow Networks (GFlowNets or GFNs) are variational inference algorithms based on Markov Decision Processes (MDP) that generate a series of terminal nodes $\mathcal{X}$ through a trajectory-based generative process \cite{bengio2023gflownet}\cite{bengio2021flow}\cite{hu2023gflownet}. They aim to align the probability of sampling instances with the values of a given reward function \cite{zhang2022generative}. This goal is fulfilled by iteratively constructing complex objects in a series of small, probabilistic steps according to the current state. GFNs treat the sampling process as a sequential decision-making problem with a learnable action policy, which provides great assistance for discovering patterns within the target distribution. To date, GFNs have showcased remarkable versatility in many fields. In bioinformatics, GFNs are instrumental in gene regulatory network modeling and demonstrate potential in unraveling complex biological systems \cite{atanackovic2024dyngfn}\cite{nguyen2023causal}, as well as in various biological sequence design efforts \cite{pmlr-v162-jain22a}. In the field of materials science, GFNs aid in automating the creation of drug-like compounds tailored to target protein pockets \cite{shen2023target}, providing assistance for biological sequence design \cite{ghari2023generative} and the exploration of other material avenues \cite{jain2023gflownets}. They have accelerated the discovery of novel materials by efficiently probing high-dimensional design spaces \cite{hernandezgarcia:hal-04407903} and have contributed to the generation of alternatives to toxic per- and poly-fluoroalkyl substances (PFAS) used in various industries \cite{soares2023a}, as well as other drug-like molecules \cite{volokhova2023towards}. Beyond biology and chemistry, GFNs have also made contributions to other domains \cite{pmlr-v202-jain23a}\cite{deleu2022bayesian}\cite{zhang2023robust}\cite{zhang2024let}.

GFNs draw inspiration from reinforcement learning (RL), with the unique advantage of sampling based on an unnormalized density specified by the policy, also known as the reward function. This makes it distinguished from other generative models trained to emulate distributions provided by training sets, which often fail to yield samples with high returns as GFNs do. Although GFNs introduce novel selection strategies, they do not address the issue regarding how to effectively leverage Markov flows to enhance exploration efficiency. Instead, its pursuit of a broad exploration space, i.e., the diversity of generated samples, significantly lowers training efficiency, hindering the acquisition of samples with higher rewards and better quality. LS-GFN \cite{kim2024local} recognized a similar issue and attempted to address it, but it was limited to local searches at the end of the sampling process. This made greedy sampling results have a low correlation to the reward values of the target's unnormalized density function, which, compared to the original GFN, unilaterally improved the reward value distribution of the sampling results.

\begin{figure}
    \centering
    \includegraphics[width=0.7\textwidth]{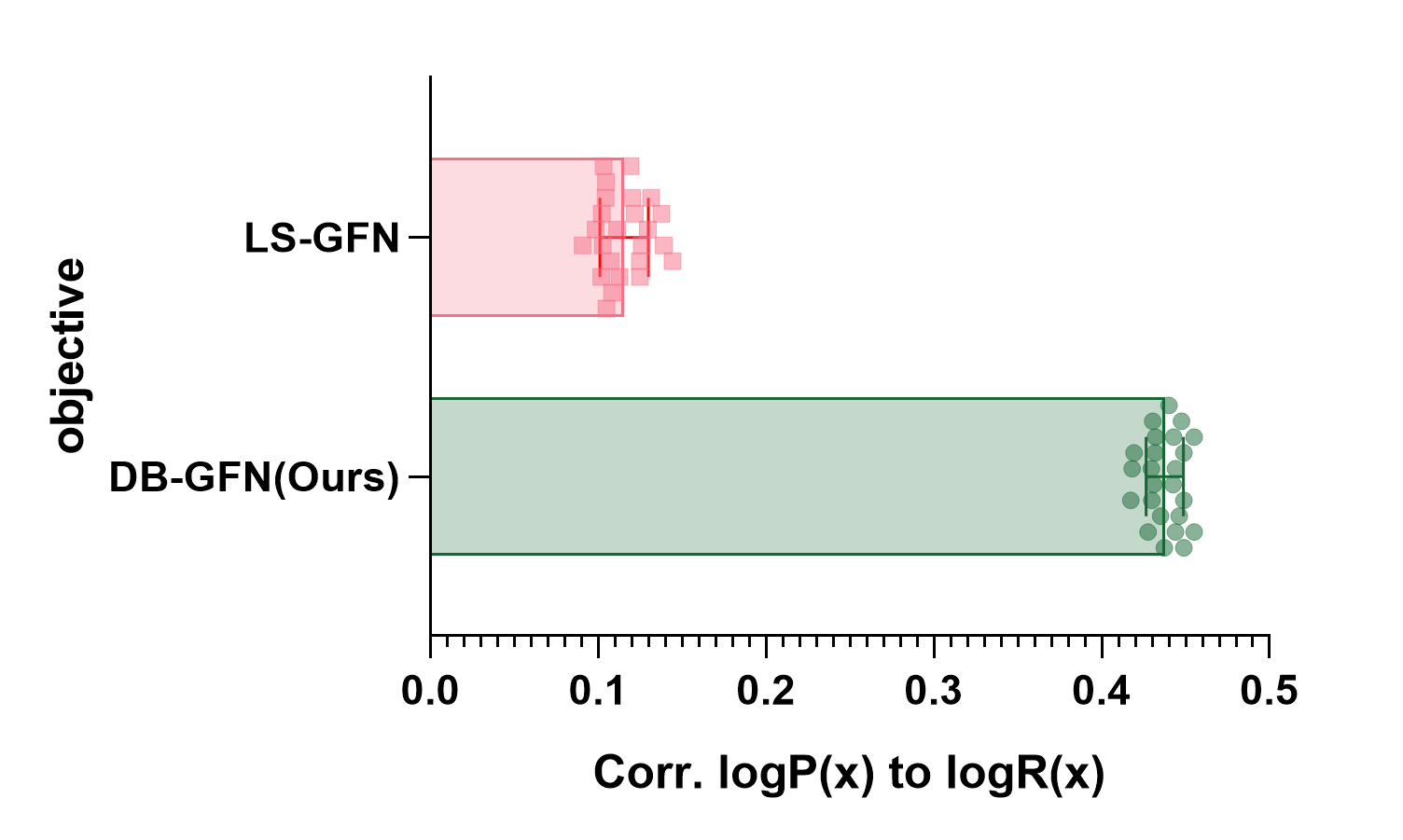}
    \caption{A motivating experiment on TFbind8, a common nucleotide generation task for the performance assessment of GFNs. Here, the Pearson correlations between reward values and sampling probabilities are displayed, where $\log{p(x)}$ is the log-likelihood for forward sampling $P_F$. The results show a significant advancement in fitting the reward function with DB-GFN over LS-GFN, with all selection strategies of DB-GFN demonstrating about a fourfold increase in correlation compared to LS-GFN.}
    \label{fig:chart of corr. Px and Rx}
\end{figure}

The DB-GFN proposed in this paper adopts a global dynamic backtracking approach. Compared with LS-GFN, DB-GFN can obtain generation results with better quality and higher rewards, evidenced by greater accuracy; also, it can avoid the low correlation caused by greedy outcomes, thereby producing results more closely aligned with the target distribution, which is an essential advantage of GFN over RL algorithms. Figure \ref{fig:chart of corr. Px and Rx} shows the correlation between reward values and sampling probabilities, which represents the likelihood of transition from initial states $s_0$ to a final state $s_n$ or termination at $x$. It is noteworthy that this correlation is one of the fundamental aspects of the GFN theory, where a perfect GFN model should exhibit a high correlation, indicating a good fit to the reward function, as demonstrated in equation \ref{eq:F eq R}.\\
\textbf{The core contribution of this study} is the introduction of a novel and implementable Markov flow learning strategy, which complements many existing improvements in the current GFN domain. DB-GFN achieves synergistic effects with other strategies and collectively improves the overall performance of GFNs, which is crucial for enhancing the scalability of this field. Through a wide range of experiments, such as biological sequence design and molecular optimization, DB-GFN demonstrates superior exploratory performance to traditional RL methods and established GFN benchmarks, and it accelerates the training convergence process. Particularly, DB-GFN achieves significant achievements in enhancing the rewards and quantity of sampling results. Moreover, it has advantages in the uniqueness of sampling results and the performance of the best batch of samples, which is of great help to practical biochemical production. 
\section{Related Work}
\label{related_work}
\paragraph{The credit assignment in GFNs} GFNs aim to ensure that the likelihood of reaching a particular termination state is proportional to its reward, rather than maximizing the reward value alone, thereby enhancing the diversity of generated outcomes. The strategies for achieving this objective include the flow matching objective \cite{bengio2021flow}, which aims to equate the incoming flow with the outgoing flow in the Markov chain, the detailed balance objective \cite{bengio2023gflownet}, which aims to satisfy the detailed balance constraint, and the trajectory balance objective \cite{NEURIPS2022_27b51bac}, which seeks to establish the trajectory balance constraint defined within the probability flow.
\paragraph{The advances and extensions of GFNs} Recent extensions of GFNs include integrating Energy-Based Models (EBMs) to enhance the generation of discrete objects \cite{ekbote2022consistent}, combining with deep learning methods to allow for more efficient adaptation to downstream tasks through pre-training, decomposing the learning process into manageable short segments of trajectories by parameterizing an additional ``flow function'' \cite{zhang2023diffusion}, enabling models to share the same neural network parameters across different temperature conditions while directly adjusting the log-odds of the policy based on the temperature value \cite{kim2023learning}, and other improvements related to the structural adjustment of Markov flows \cite{ma2023baking}\cite{rector2023thompson}.
\section{Preliminaries}
\label{preliminaries}
This paper mainly utilizes the notational conventions in \cite{bengio2023gflownet} to define some of the terminologies in GFN theory. GFNs are generative models for producing compositional objects such as chemical substances and biological proteins, denoted by $x\in\mathcal{X}$. These compositional objects represent the terminal points of a MDP. These terminal points are called \textit{terminal states}, which, when input into the reward function $\mathbb{R}$, yield a reward value $R(x)\ge0$ for the respective terminal states. GFN utilizes amortized inference algorithms to overcome the probability distribution challenges in the MDP. Extensive research has been conducted on the exploration of the GFN generation process, including the generation of \textit{terminal states} and extending into continuous domains, as demonstrated in the recent work \cite{lahlou2023theory}. Learning signals can be obtained not just after receiving rewards at \textit{terminal states} but throughout the generation process \cite{zhang2023diffusion}. Dynamic Backtracking GFN (DB-GFN) is orthogonal to these approaches. It can be used in these new lines of inquiry, although this paper focuses on discrete GFNs that receive rewards at \textit{terminal states}.

Let $\mathcal{G}=(\mathcal{S,A})$ be the directed acyclic graph (DAG) produced in the Markov decision process, regarded as a collection of \textbf{trajectory flows} $(\mathcal{T})$. There exists a set of states $s\in\mathcal{S}$, among which the initial state $s_0\in\mathcal{S}$, also known as the source state, is unique as it is the only state without incoming edges. Each edge, referred to as an action $(s\rightarrow s^\prime)\in \mathcal{A} \subseteq \mathcal{S}\times\mathcal{S}$, signifies the decision-making step from the initial state to the terminal state, which is essential for shaping the intended GFN structure. A complete trajectory $(\tau)$, consisting of a series of states $s$ and directed edges, is denoted as $\tau=(s_0\rightarrow s_1 \rightarrow ... \rightarrow s_n)$, where $s_0$ represents the starting point for all $\tau$ and leading to a set of terminal states $s_n = x \in \mathcal{X}\subseteq \mathcal{S}$, identified as nodes with no outgoing edges in relation to the initial state. The correlation between $\tau$ and $s_n$ facilitates the definition of a non-negative, unnormalized density function $F(\tau): \mathcal{T}\rightarrow \mathbb{R}$. Therefore, for a state, we have $F(s)=\Sigma_{\tau \in \mathcal{T}:s \in \tau}F(\tau)$, and for an action, $F(s\rightarrow s^\prime)= \Sigma_{\tau \in \mathcal{T}:(s\rightarrow s^\prime)\in\tau}F(\tau)$. Moreover, for the transition $s\rightarrow s^\prime$, the pivotal forward policy $P_F(s^\prime\mid s)$ and $P_F(\tau)$ and the backward policy $P_B(s\mid s^\prime)$ and $P_B(\tau)$ can be formulated. Then, for $s\rightarrow s^\prime$, the crucial forward policy $P_F(s^\prime\mid s)$ and $P_F(\tau)$ and the backward policy $P_B(s\mid s^\prime)$ and $P_B(\tau)$ can be defined.
The forward policy probability $P_F$ for a given trajectory $\tau$ is defined as:
\begin{equation}\label{eq:P_F(tau)}
 P_F(\tau=(s_0\rightarrow s_1 \rightarrow ... \rightarrow s_n)):=\prod_{t=0}^{n-1}P_F(s_{t+1}\mid s_t)
\end{equation}
Here, the conditional probability $P_F(s_{t+1}\mid s_t)$ is determined by the ratio of $F(s_t\rightarrow s_{t+1})$ to $F(s)$:
\begin{equation}\label{eq:PFtoF/F}
    P_F(s_{t+1}\mid s_t):=\frac{F(s_{t}\rightarrow s_{t+1})}{F(s)}
\end{equation}
Similarly, the trajectory probability for the backward policy $P_B$ is expressed as:
\begin{equation}\label{eq:P_B(tau)}
 P_B(\tau=(s_0\rightarrow s_1 \rightarrow ... \rightarrow s_n)):=\prod_{t=0}^{n-1}P_B(s_{t}\mid s_{t+1})
\end{equation}
where the conditional probability $P_B(s_t\mid s_{t+1})$ is calculated using the ratio of $F(s_{t}\rightarrow s_{t+1})$ to $F(s_{t+1})$:
\begin{equation}\label{eq:PBtoF/F}
    P_B(s_{t}\mid s_{t+1}):=\frac{F(s_{t}\rightarrow s_{t+1})}{F(s_{t+1})}
\end{equation}
\section{Dynamic Backtracking in GFN (DB-GFN)}
\label{DB-GFN}
\textbf{Overview:} As indicated by equation \ref{eq:F eq R}, this objective is the core of GFN, but it is a double-edged sword. It enables GFNs to explore a broader sampling space based on the distribution of the reward function $R(x)$, thus discovering high-quality compositions that are more likely to occur in real life. However, when the targeted sampling space is huge, GFNs may fail to focus on generating higher-quality samples. Instead, they could be affected by the noise within the space, struggling to find better samples, or even getting lost in high-dimensional spaces, which is undesirable.

This paper believes that this phenomenon arises from the unidirectional progression in the Markov flow generation process, where states can only move forward. This prevents self-correction upon encountering suboptimal outcomes and inhibits the possibility of retracing to explore alternative pathways. This is analogous to playing a game of chess, where if we find ourselves at a disadvantage, we might choose to withdraw the pieces, and the worse the situation, the more we hope to withdraw more pieces to achieve a better outcome. It is noteworthy that, in our Markov flow dynamic backtracking, the self-correction process occurs before the training of the model and the adjustment of parameters. This helps bypass local optima that may arise during parameter optimization, leading to an improved model.

\begin{figure}
    \centering
    \includegraphics[width=0.9\textwidth]{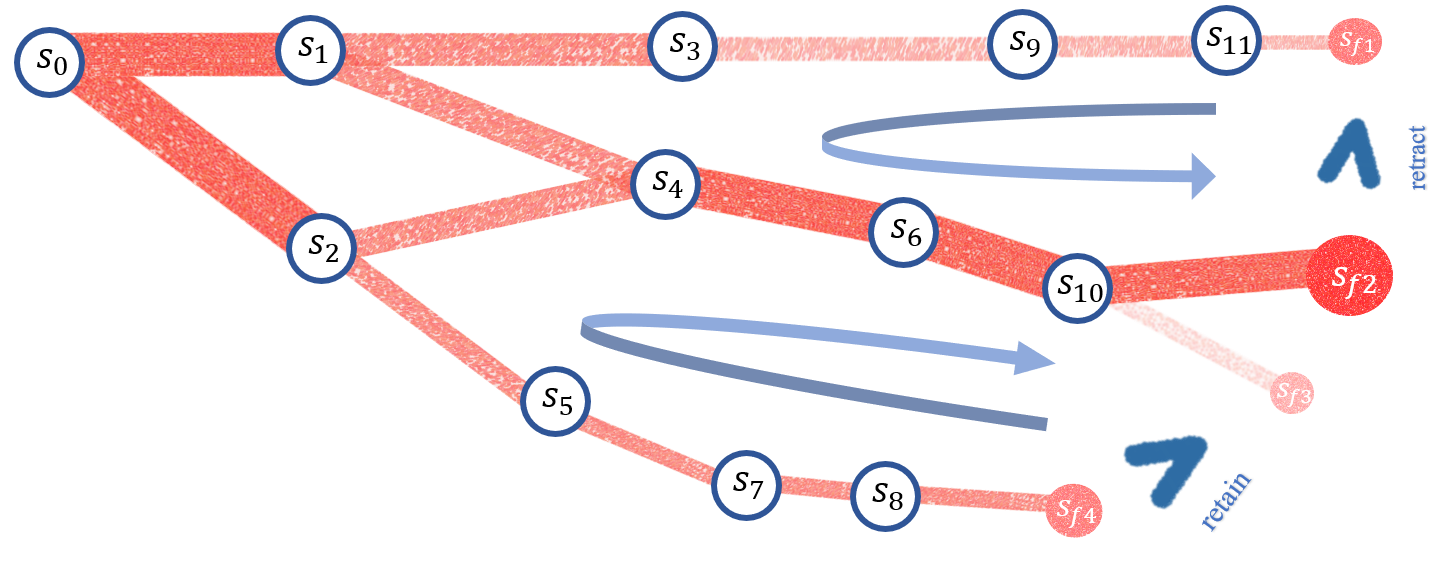}
    \caption{The structure of DB-GFN is illustrated with terminal states $s_{f\{1..4\}}$, where $s_{f1}$ and $s_{f4}$ represent the initially generated terminal states. The size and shading of states indicate the magnitude of rewards, and the transparency and thickness of flows represent the probability magnitude. In this illustration, $s_{f4}$ with a smaller reward undergoes more backtracking, in this case, to $s_2$, and then transitions to new terminal states. After applying the Reward-Choose selection algorithm, one of three such algorithms considered, the newly generated terminal state $s_{f3}$ is less favorable than the original $s_{f4}$, so the original trajectory is kept. Similarly, $s_{f1}$ has a larger reward than $s_{f4}$ and undergoes fewer backtracking steps. The new outcome $s_{f2}$, produced after backtracking, is preferred in the selection algorithms, leading to an update and the choice of a new trajectory ($s_0\rightarrow...\rightarrow s_2\rightarrow...\rightarrow s_{f2}$) over the original one ($s_0\rightarrow...\rightarrow s_2\rightarrow...\rightarrow s_{f1}$).}
    \label{fig:DBGFN}
\end{figure}

To sum up, the workflow of DB-GFN is as follows:\\
\textbf{Forward Propagation:} Initially, GFN performs forward propagation from the starting state $s_0$, making forward decisions to sample a batch of terminal outcomes, and an initial probability propagation flow is generated, as shown in Figure \ref{fig:DBGFN}.\\
\textbf{Dynamic Backtracking:} Then, a random selection is performed to give some nodes the opportunity for dynamic backtracking. Each terminal state that has the chance to backtrack determines the number of steps that is inversely related to its reward value globally, as shown in equation \ref{eq:step by reward}, and then backtracking occurs.\\
\textbf{Partial Forward Propagation:} Nodes that have been backtracked undergo forward propagation again to find more efficient pathways.\\
\textbf{Compare and Choose:} After comparing the new sampling outcomes with the original ones, the network chooses to retain the more valuable results to escape local optima, guided by the comparison algorithms in equations \ref{eq:choose rewarder}, \ref{eq:choose pearson}, and \ref{eq: choose MH}.\\
It should be noted that the above DB-GFN process only involves the generation of Markov flows and does not include the subsequent amortized sampling from the corresponding probability density function on a space of compositional objects $\mathcal{X}$. This indicates that DB-GFN can be used in conjunction with other GFN improvements in the amortization aspect.
\subsection{Dynamic Backtracking}
\subsubsection{Dynamic Regret}
To consider the potential greed effect resulting from the ``undo'' option deliberated for each sample, particularly when it could lead to an undue focus on target features and a decrease in the number of samples in smaller-scale outcomes, this paper implements a hyperparameter $Tg$ for the selection of dynamic sampling probability. The regret probability $\mathbb{B}$ is defined as $\mathbb{B}=(1-e^{-Tg})$, which enables a dynamic effect of regret.
\subsubsection{Dynamic Step}
To address the issue of traditional GFNs potentially obtaining poor results and falling into local optima during model training, this paper uses an algorithm to decide on multi-step backtracking based on the quality of the current reward value. Specifically, the number of actual backtracking steps $\mathbb{S}$, as well as the maximum number of steps $S_m$ and minimum number of steps $S_l$, and the upper bound $T_m$ and lower bound $T_l$ for backtracking reward evaluation, all follow the formula outlined in \ref{eq:step by reward}.
\begin{equation}
\label{eq:step by reward}
\mathbb{S} = 
\left\{
\begin{array}{lll}
S_m & \text{if } T_l \leq R,\\
\left\lfloor S_m-\frac{(S_m-S_l)\times (R-T_l)}{T_m-T_l}\right \rfloor& \text{if } T_l < R <T_m,\\
S_l &\text{if } R \leq T_m.
\end{array}
\right.
\end{equation}
With this approach, the number of backtracking steps is dynamically selected based on the current sampling situation, which enables us to find better sampling outcomes within a large sampling space.
\subsection{Dynamic Choose}
Drawing inspiration from \cite{kim2024local}, three methods are devised to assess whether the current situation is disadvantageous. The new path after dynamic backtracking is denoted as $\tau^\prime$, and the original path is denoted as $\tau$, with $\mathbb{C}$ being the probability of accepting the backtracked result.
\paragraph{Reward-Choose:} The trajectory that yields the highest reward during the sampling process is selected, as indicated by the selector in \ref{eq:choose rewarder}.
\begin{equation}\label{eq:choose rewarder}
    \mathbb{C}(\tau^\prime\mid \tau) = {\mathbbm{1}}_{(R(\tau^\prime)> R(\tau))}
\end{equation}
\paragraph{Pearson-Choose:} After dynamically backtracking a batch of samples, the Pearson correlation coefficient is utilized to compare the degree to which $P(x)\propto R(x)$, thus assessing the quality of the sampling.
\begin{equation}
    r = \frac{\sum [P(x_i) - \overline{P(x)}][R(x_i) - \overline{R(x)}]}{\sqrt{\sum [P(x_i) - \overline{P(x)}]^2} \sqrt{\sum [R(x_i) - \overline{R(x)}]^2}}
\end{equation}
\begin{equation}\label{eq:choose pearson}
    \mathbb{C}(\tau^\prime\mid \tau)={\mathbbm{1}}_{(r_{\tau^\prime}> r_{\tau})}
\end{equation}
\paragraph{MH-Choose:} To decide on accepting a new trajectory, the acceptance probability $\mathbb{C}$ is calculated, which is based on the reward of the target sampling and the probabilities along the path. The divergence point between the two paths is denoted as $S_d$.
\begin{equation}\label{eq: choose MH}
    \mathbb{C}(\tau^\prime \mid \tau)=\frac{R(x^\prime)\times P_B(x^\prime\rightarrow ... \rightarrow S_d)\times P_F(S_d\rightarrow...\rightarrow x)}{R(x)\times P_B(x\rightarrow ... \rightarrow S_d)\times P_F(S_d\rightarrow...\rightarrow x^\prime)}
\end{equation}
\section{Experiments}
This paper presents the results for six biochemical molecular tasks for comparison with LS-GFN \cite{kim2024local}, thus demonstrating our notable improvements in the GFNs. Meanwhile, comparisons are performed by taking other GFN and RL models as baselines. The comparison indicators include the number of substances generated through GFN sampling, the fit of the reward function, the average score of the sampled results, the performance of the best sampling results, and the uniqueness of the outcomes, all of which are widely recognized indicators for the validation of substances generated by GFN sampling \cite{bengio2023gflownet}\cite{pmlr-v162-jain22a}\cite{NEURIPS2022_27b51bac}\cite{hernandezgarcia:hal-04407903}. Specifically, the final score is calculated using the accuracy calculation method from \cite{shen23accuracy} as shown in \ref{eq: accuracy}, where \(p(x;\theta)\) denotes the result distribution obtained from training and \(p^\prime(x;\theta)\) represents the target distribution.
\begin{equation}\label{eq: accuracy}
    Accuracy=min\left(\frac{E_{p(x;\theta)}[R(x)]}{E_{p^\prime(x;\theta)}[R(x)]},1\right)
\end{equation}
\subsection{Baselines}
Our approach was compared with previous GFN methods and RL techniques used for molecular generation.

\textbf{GFNs:} The GFNs methods include detailed balance GFN (DB)\cite{bengio2023gflownet}, trajectory balance GFN (TB)\cite{NEURIPS2022_27b51bac}, substructure-guided trajectory balance GFN (GTB)\cite{shen23accuracy}, sub-trajectory balance GFN (SubTB)\cite{madan2023learning}, and maximum entropy GFN (MaxEntGFN)\cite{NEURIPS2022_27b51bac}.

\textbf{RL:} The RL methods include advantage actor-critic (A2C), proximal policy optimization (PPO)\cite{schulman2017proximal}, and Markov molecular sampling (MARS)\cite{xie2021mars}.

To ensure fairness during the training process, this paper strictly followed the implementation procedures described in \cite{kim2024local} and \cite{shen23accuracy}, which included existing models such as LS-GFN and GTB. This involved maintaining consistency in hyperparameter settings, such as the number of samples per training iteration, the number of local search revisions in LS-GFN, and the budget for reward evaluation. Meanwhile, to eliminate the impact of other variables, our proposed DB-GFN used the same hyperparameters as the existing models. Accuracy evaluation was performed with testing of 128 untrained online samples every 10 rounds, indicating that the DB-GFN method is only applicable to the training stage, not to the inference stage. Additionally, each experiment was carried out with three different random seeds for testing.
\subsection{Implementations}
To demonstrate the significant advantages of our model over the previous GFN tasks (including LS-GFN and other GFN models), experiments were conducted on the six tasks previously tackled by LS-GFN. The results indicate that our model not only outperformed LS-GFN and other GFN and RL models in terms of accuracy but also significantly surpassed them in the number of sampling results.
\subsubsection{Biochemical Molecules} 
The first task, \textbf{QM9}, aims to generate small molecular graphs that maximize the HOMO-LUMO gap, with a path space $\lvert\mathcal{T}\rvert$ of 940,240 and a final state space $\lvert\mathcal{X}\rvert$ of 58,765. The reward function is provided by a pre-trained agent, the MXMNet proxy \cite{zhang2020molecular}. The second task, \textbf{sEH}, aims to generate protein sequences with a high binding affinity, with a path space $\lvert\mathcal{T}\rvert$ of 1,088,391,168 and a final state space $\lvert\mathcal{X}\rvert$ of 34,012,244, where the reward function is based on the protein's binding affinity \cite{bengio2021flow}.

\begin{figure}
    \centering
    \includegraphics[width=0.9\textwidth]{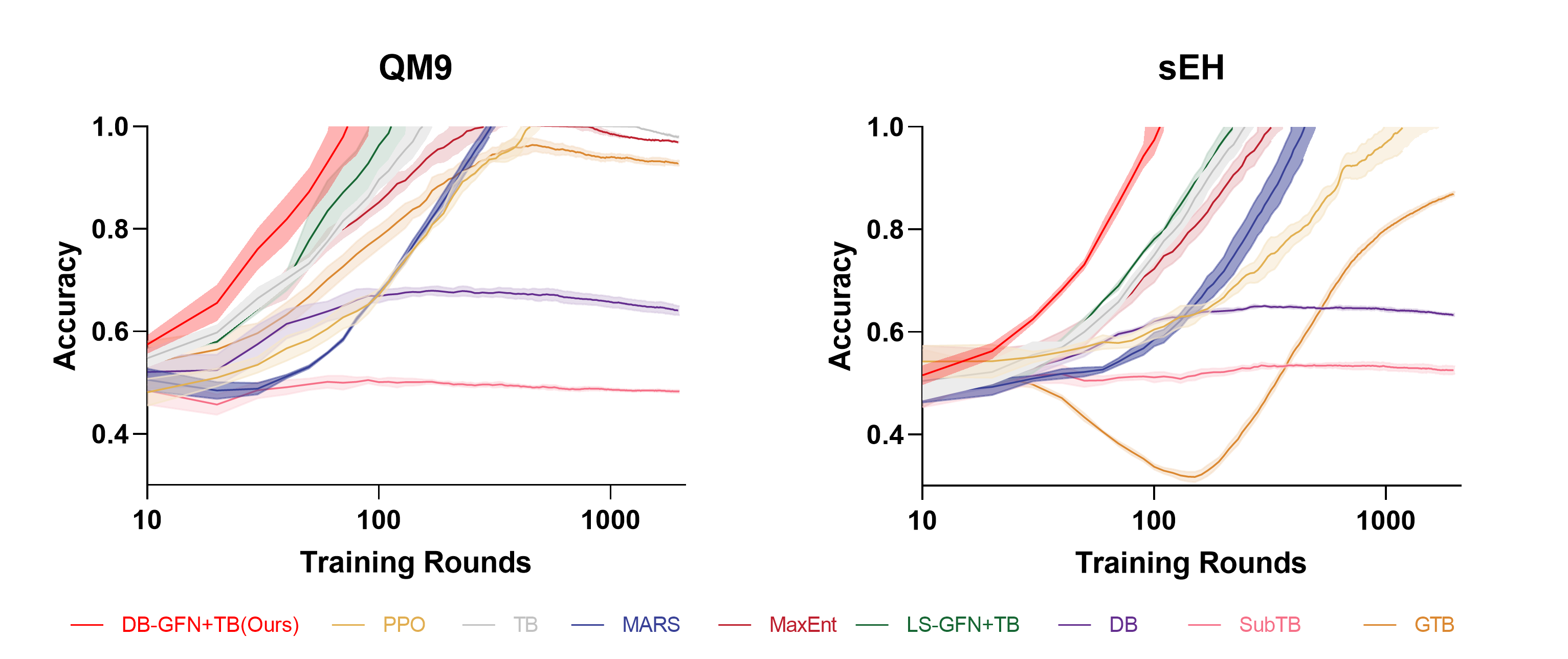}
    \caption{Accuracy of GFNs on QM9 and SEH tasks, with our DB-GFN model significantly outperforming other models.}
    \label{fig:molecule acc}
\end{figure}

For both QM9 and sEH tasks, the number of training rounds was set to 2000. As demonstrated in Figure \ref{fig:molecule acc}, our model, DB-GFN, not only achieves the highest accuracy more quickly but also exhibits greater stability in the improvement of accuracy. Table \ref{tab:molecule num} presents the great improvement of our model in discovering a large number of high-scoring sampled biochemical molecules. It is noteworthy that ``discovered sampled molecules'' refers to those molecular samples that meet certain criteria and exceed a specified reward threshold.
\begin{table}
    \centering
    \begin{adjustbox}{width=\textwidth}
    \begin{tabular}{ccccccccccc}
        \toprule
         Task&  DB-GFN(Ours) &  LS-GFN &  TB &  MaxEnt &  MARS &  GTB &  DB &  subTB & PPO & A2C\\
         \midrule
         QM9($\uparrow$)& \boldmath{$714\pm4$} &  $649\pm23$& $536\pm16$ & $524\pm3$ & $373\pm24$ & $483\pm18$ & $165\pm3$ & $115\pm5$ &  $396\pm72$ & $67\pm21$\\
         SEH($\uparrow$)& \boldmath{$2623\pm54$} & $1521\pm165$ & $342\pm7$ & $302\pm18$ & $1955\pm477$ & $91\pm5$ & $22\pm4$ & $16\pm5$ & $200\pm56$ & $2205\pm990$ \\
         \bottomrule
    \end{tabular}
    \end{adjustbox}
    \caption{The number of modes discovered on QM9 and sEH tasks. The generality of the results is illustrated through the average and standard deviation across three random seeds}
    \label{tab:molecule num}
\end{table}
\subsubsection{Genetic Material Sequences}
The first task, \textbf{RNA-Binding}, involves three experiments of RNA molecule generation with distinct transcriptional targets. It aims to generate an RNA sequence composed of 14 nucleotides, with a path space $\lvert\mathcal{T}\rvert$ of 2,199,023,255,552 and a final state space $\lvert\mathcal{X}\rvert$ of 268,435,456. Meanwhile, there are three distinct transcriptional targets from L14-RNA1 to L14-RNA3, each associated with a different reward function \cite{sinai2021adalead}. The second task, \textbf{TFBind8}, aims to generate an 8-nucleotide long sequence, with a path space $\lvert\mathcal{T}\rvert$ of 8,388,608 and a final state space $\lvert\mathcal{X}\rvert$ of 65,536, where the reward function is the binding affinity between DNA and human transcription factors \cite{trabucco2022design}.

\begin{figure}
    \centering
    \includegraphics[width=1\textwidth]{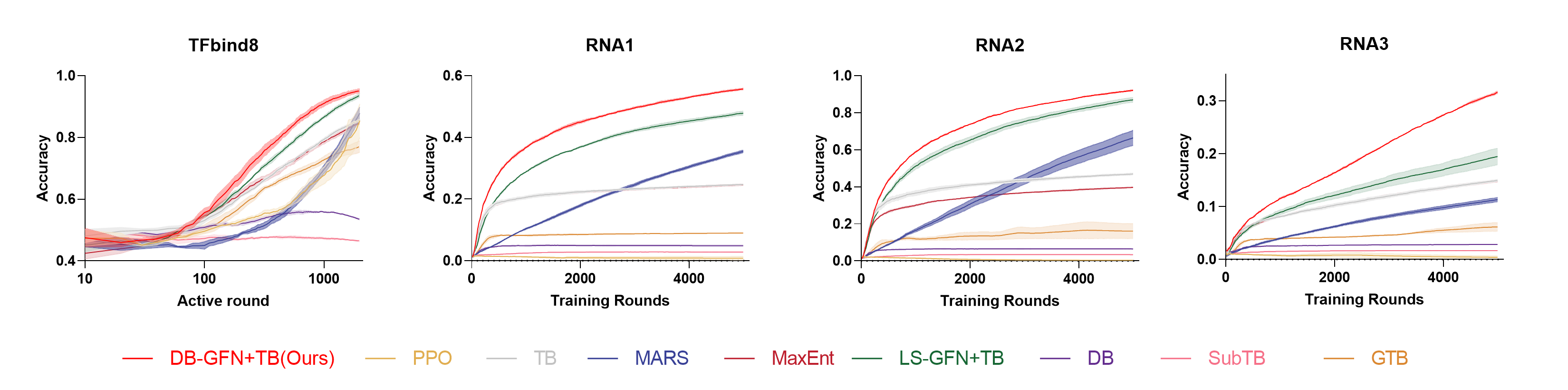}
    \caption{Accuracy comparison of GFNs on the TFbind8 and RNA1-3 tasks, where our DB-GFN model significantly outperforms the other models}
    \label{fig:drna acc}
\end{figure}

For the TFBind8 task, 2000 training rounds were performed, while for the three RNA molecule generation tasks, 5000 training rounds were performed. The results demonstrate that DB-GFN consistently outperforms other models in the sampling outcomes for these four genetic material sequence tasks. As illustrated in Figure \ref{fig:drna acc}, DB-GFN achieves faster convergence and higher accuracy in both the TFBind8 and the three RNA molecule generation tasks. Notably, DB-GFN significantly outperforms other models from the onset of the training in the RNA tasks. Additionally, Table \ref{tab:D/RNA num} suggests that DB-GFN is not only good at finding high-scoring samples but also capable of discovering a larger number of viable sampling specimens, particularly evident in the RNA tasks where our model showed an overwhelming advantage without any hyperparameter optimization.
\begin{table}
    \centering
    \begin{adjustbox}{width=\textwidth}
    \begin{tabular}{ccccccccccc}
        \toprule
         Task&  DB-GFN(Ours) &  LS-GFN &  TB &  MaxEnt &  MARS &  GTB &  DB &  subTB & PPO & A2C\\
         \midrule
         TFbind8($\uparrow$)& \boldmath{$278\pm4$} &  $270\pm4$& $266\pm4$ & $261\pm6$ & $206\pm6$ & $234\pm10$ & $156\pm13$ & $121\pm5$ &  $230\pm12$ & $19\pm4$\\
         RNA1($\uparrow$)& \boldmath{$19743\pm 132$} & $16\pm 2$ & $5\pm 2$ & $9442\pm 14$ & $9\pm 2$ & $2513\pm27 $ & $1223\pm 120$ & $590\pm 23$ & $26\pm 21$ & $3\pm 1$ \\
         RNA2($\uparrow$)& \boldmath{$23197\pm 187$} & $13\pm 4$ & $3\pm 2$ & $12028\pm 158$ & $5\pm 3$ & $4257\pm 2056$ & $1343\pm 107$ & $577\pm 2$ & $0\pm 0$ & $0\pm 0$ \\
         RNA3($\uparrow$)& \boldmath{$13369\pm 44$} & $5\pm 2$ & $4862\pm 3434$ & $7308\pm 85$ & $4055\pm 2837$ & $ 2832\pm713 $ & $1019\pm 22$ & $518\pm 18$ & $26\pm 37$ & $0\pm 0$ \\
         \bottomrule
    \end{tabular}
    \end{adjustbox}
    \caption{The number of modes discovered on TFbind8 and RNA tasks}
    \label{tab:D/RNA num}
\end{table}
\subsubsection{Comprehensive Comparison}
\begin{figure}
    \centering
    \includegraphics[width=0.9\textwidth]{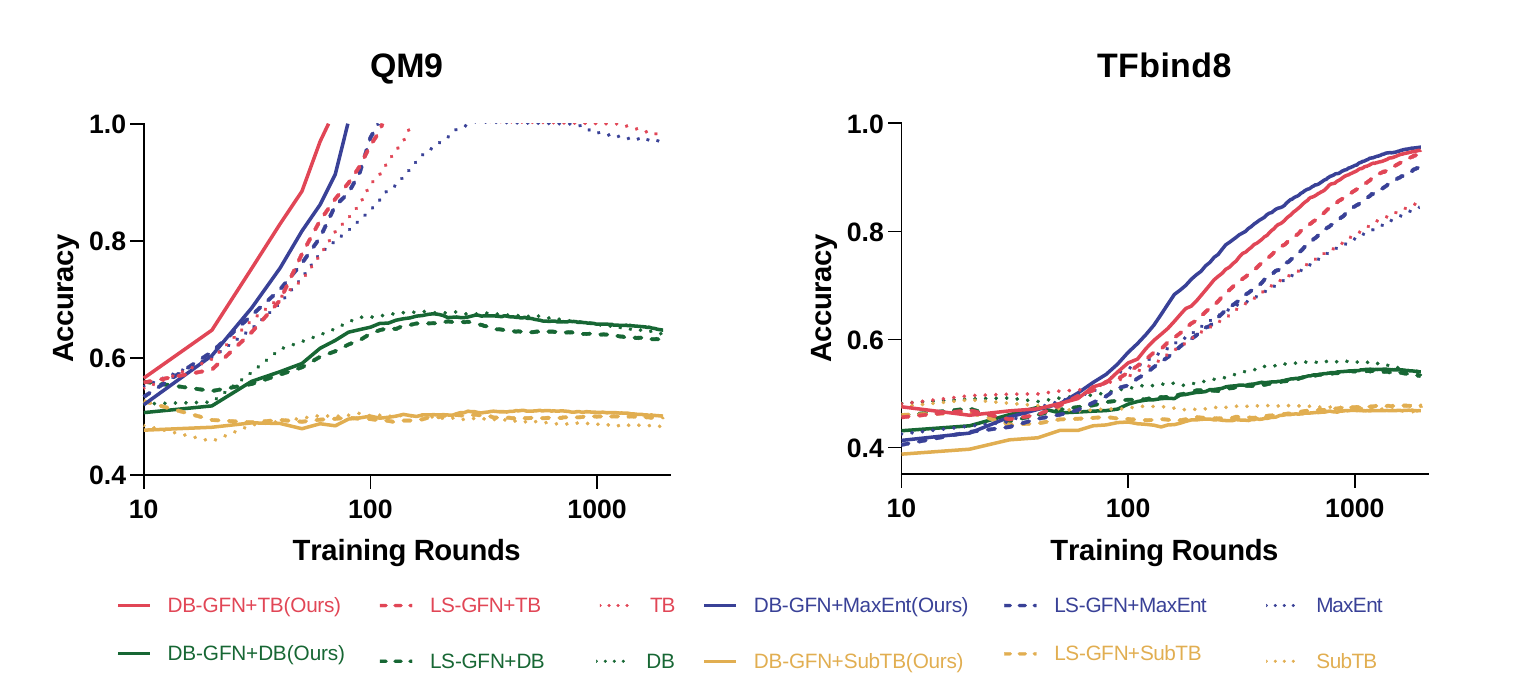}
    \caption{Accuracy for the TFbind8 and RNA tasks with DB-GFN represented by solid lines. Notably, integrating DB-GFN into the same models (indicated by matching colors) consistently yields better performance than both the original baselines and those augmented with LS-GFN, as demonstrated by the remarkable results indicated by the solid lines .}
    \label{fig:old-ls-db acc}
\end{figure}

\begin{figure}
    \centering
    \includegraphics[width=0.9\linewidth]{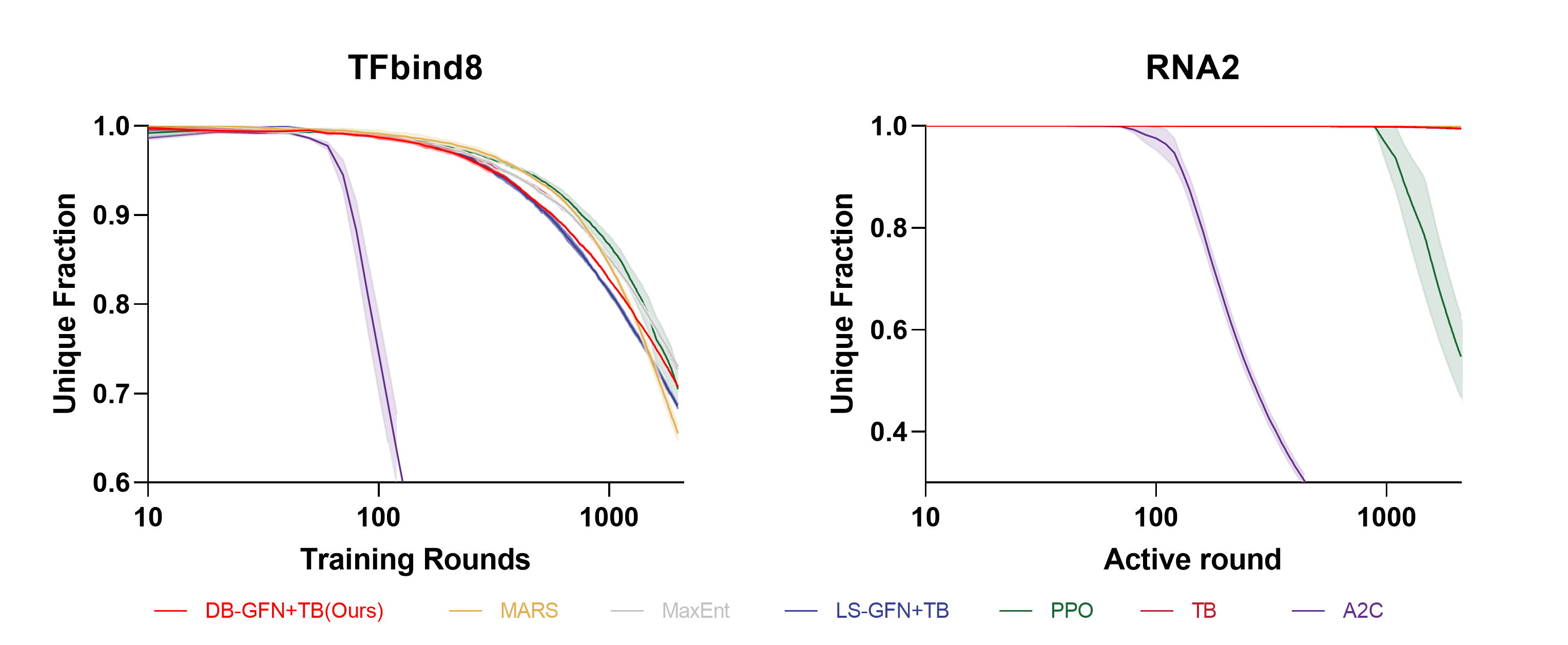}
    \caption{The uniqueness score of sampling outcomes for the TFbind8 task, with 1 being the highest score, indicating that all high-reward samples found are unique and non-repeating.}
    \label{fig:unique}
\end{figure}

Firstly, this study conducted an orthogonal comparison of our DB-GFN against other GFN models including subTB, TB, DB, and MaxEnt across two tasks: the molecular task QM9 and the genetic sequence task TFbind8. To demonstrate the improvements of DB-GFN not only in its original version but also relative to the LS-GFN baseline, LS-GFN baselines were set for each model. As shown in Figure \ref{fig:old-ls-db acc}, DB-GFN significantly improves performance on TB and MaxEnt tasks. Similarly, as listed in Table \ref{tab:old-ls-db num}, in terms of the number of modes, DB-GFN shows considerable improvement on the DB and TB tasks and performs well on the MaxEnt and subTB tasks. This underscores the generalizability and versatility of DB-GFN, which can be applied to existing and even future GFN models to achieve performance improvements on various model foundations.
\begin{table}
    \centering
    \begin{adjustbox}{width=\textwidth}
    \begin{tabular}{ccccccc}
        \toprule
         Task&  DB &  DB+LS-GFN & DB+DB-GFN(Ours)($\uparrow\uparrow$)  &  MaxEnt &  MaxEnt+LS-GFN &  MaxEnt+DB-GFN(Ours)($\uparrow$) \\
         \midrule
         TFbind8& $156\pm 13$ &  $145\pm 14$& \boldmath{$160\pm 10$} & $261\pm 6$ & $264\pm 3$ & \boldmath{$277\pm 2$} \\
         QM9& $165\pm 3$ &  $156\pm 15$& \boldmath{$180\pm 7$} & $524\pm 3$ & \boldmath{$728\pm 11$} & $699\pm 14$ \\
         \bottomrule
         \toprule
         Task&  subTB &  subTB+LS-GFN & subTB+DB-GFN(Ours) & TB & TB+LS-GFN & TB+DB-GFN(Ours)($\uparrow\uparrow$)\\
         \midrule
         TFbind8& $121\pm 5$ &  \boldmath{$129\pm 9$}& $122\pm 7$ & $266\pm 4$ & $270\pm 4$ &\boldmath{ $277\pm 2$} \\
         QM9& \boldmath{$115\pm 7$} &  $103\pm 7$& $111\pm 6$ & $536\pm 16$ & $691\pm 7$ & \boldmath{$714\pm 4$} \\
         \bottomrule
    \end{tabular}
    \end{adjustbox}
    \caption{The number of modes discovered on TFbind8 and RNA tasks. Using the DB-GFN approach on the DB, TB, and MaxEnt GFN models yields significant performance improvements.}
    \label{tab:old-ls-db num}
\end{table}
Furthermore, from Figure \ref{fig:chart of corr. Px and Rx}, it can be seen that our DB-GFN model demonstrates a nearly fourfold improvement in Pearson correlation compared to LS-GFN, indicating a superior fitting to the reward function; also, the practicality of DB-GFN in actual biochemical molecular experiments is evident by comparing the uniqueness of sampling outcomes displayed in Figure \ref{fig:unique} and the performance of the top sampling outcomes listed in Table \ref{tab:top accuracy}. Figure \ref{fig:unique} compares the uniqueness of sampling results for the TFbind8 task among DB-GFN, LS-GFN, and the four GFN models TB, MaxEnt, alongside all RL models, where uniqueness is calculated as the ratio of the number of effective high-reward samples to the total number of high-reward samples. The results indicate a commendable uniqueness for DB-GFN. Table \ref{tab:top accuracy} presents the average reward of the top 100 samples by reward value for the L14-RNA2 and TFbind8 tasks, highlighting only the top six models with relatively higher overall average rewards. Obviously, DB-GFN achieves the best reward performance.
\begin{table}
    \centering
    \begin{adjustbox}{width=1\textwidth}
    \begin{tabular}{ccccccc}
    \toprule
        Task & MaxEnt & MARS & GTB & DB-GFN(Ours) & LS-GFN & TB \\
    \midrule
        L14-RNA2 & $0.895\pm 0.0031$ & $0.919\pm 0.0061$ & $0.820\pm 0.0230$ & \boldmath{$0.952\pm 0.0017$} & $0.948\pm 0.0006$ & $0.898\pm 0.0041$ \\
        TFbind8 & $0.982\pm 0.0004$ & $0.979\pm 0.0004$ & $0.982\pm 0.0010$ & \boldmath{$0.983\pm 0.0001$} & $0.983\pm 0.0001$ & $0.982\pm 0.0003$ \\
    \bottomrule
    \end{tabular}
    \end{adjustbox}
    \caption{The average performance of the top 100 sampled outcomes by reward value in the L14-RNA2 and TFbind8 tasks. These six models have comparatively better overall performance. Obviously, DB-GFN stands out. The table presents the average values and standard deviations from three random seeds for each model.}
    \label{tab:top accuracy}
\end{table}
\section{Conclusion}
This paper introduces a novel GFN methodology that implements a dynamic backtracking mechanism based on reward values to obtain improved training outcomes. Our empirical evidence demonstrates that this method not only increases the number of sampled outcomes produced by GFN models but also enhances the exploration of the sampling space. By dynamically backtracking based on previous exploration experiences, superior sampling results are obtained. This significantly improves training efficiency and mitigates the risk of being lost in the search space due to a one-way forward approach that could lead to loss of experience. Specifically, in our six biochemical generation tasks, the DB-GFN model consistently achieved faster convergence speeds and higher convergence values than eight baselines, including GFN models like LS-GFN and TB, as well as RL models such as PPO. An exciting and noteworthy point is that our GFN is orthogonal to many well-improved GFN models, suggesting that DB-GFN is a vital tool for future improvements in GFN networks. DB-GFN can be integrated with other concepts to realize even more efficient GFN networks.
\bibliographystyle{unsrtnat}
\bibliography{db-gfn}

\begin{thebibliography}{32}
\providecommand{\natexlab}[1]{#1}
\providecommand{\url}[1]{\texttt{#1}}
\expandafter\ifx\csname urlstyle\endcsname\relax
  \providecommand{\doi}[1]{doi: #1}\else
  \providecommand{\doi}{doi: \begingroup \urlstyle{rm}\Url}\fi

\bibitem[Bengio et~al.(2023)Bengio, Lahlou, Deleu, Hu, Tiwari, and Bengio]{bengio2023gflownet}
Yoshua Bengio, Salem Lahlou, Tristan Deleu, Edward~J Hu, Mo~Tiwari, and Emmanuel Bengio.
\newblock Gflownet foundations.
\newblock \emph{Journal of Machine Learning Research}, 24\penalty0 (210):\penalty0 1--55, 2023.

\bibitem[Bengio et~al.(2021)Bengio, Jain, Korablyov, Precup, and Bengio]{bengio2021flow}
Emmanuel Bengio, Moksh Jain, Maksym Korablyov, Doina Precup, and Yoshua Bengio.
\newblock Flow network based generative models for non-iterative diverse candidate generation.
\newblock \emph{Advances in Neural Information Processing Systems}, 34:\penalty0 27381--27394, 2021.

\bibitem[Hu et~al.(2023)Hu, Malkin, Jain, Everett, Graikos, and Bengio]{hu2023gflownet}
Edward~J Hu, Nikolay Malkin, Moksh Jain, Katie~E Everett, Alexandros Graikos, and Yoshua Bengio.
\newblock Gflownet-em for learning compositional latent variable models.
\newblock In \emph{International Conference on Machine Learning}, pages 13528--13549. PMLR, 2023.

\bibitem[Zhang et~al.(2022)Zhang, Malkin, Liu, Volokhova, Courville, and Bengio]{zhang2022generative}
Dinghuai Zhang, Nikolay Malkin, Zhen Liu, Alexandra Volokhova, Aaron Courville, and Yoshua Bengio.
\newblock Generative flow networks for discrete probabilistic modeling.
\newblock In \emph{International Conference on Machine Learning}, pages 26412--26428. PMLR, 2022.

\bibitem[Atanackovic et~al.(2024)Atanackovic, Tong, Wang, Lee, Bengio, and Hartford]{atanackovic2024dyngfn}
Lazar Atanackovic, Alexander Tong, Bo~Wang, Leo~J Lee, Yoshua Bengio, and Jason~S Hartford.
\newblock Dyngfn: Towards bayesian inference of gene regulatory networks with gflownets.
\newblock \emph{Advances in Neural Information Processing Systems}, 36, 2024.

\bibitem[Nguyen et~al.(2023)Nguyen, Tong, Madan, Bengio, and Liu]{nguyen2023causal}
Trang Nguyen, Alexander Tong, Kanika Madan, Yoshua Bengio, and Dianbo Liu.
\newblock Causal inference in gene regulatory networks with gflownet: Towards scalability in large systems.
\newblock \emph{arXiv preprint arXiv:2310.03579}, 2023.

\bibitem[Jain et~al.(2022)Jain, Bengio, Hernandez-Garcia, Rector-Brooks, Dossou, Ekbote, Fu, Zhang, Kilgour, Zhang, et~al.]{pmlr-v162-jain22a}
Moksh Jain, Emmanuel Bengio, Alex Hernandez-Garcia, Jarrid Rector-Brooks, Bonaventure~FP Dossou, Chanakya~Ajit Ekbote, Jie Fu, Tianyu Zhang, Michael Kilgour, Dinghuai Zhang, et~al.
\newblock Biological sequence design with gflownets.
\newblock In \emph{International Conference on Machine Learning}, pages 9786--9801. PMLR, 2022.

\bibitem[Shen et~al.(2023{\natexlab{a}})Shen, Pandey, Smith, Cherkasov, and Ester]{shen2023target}
Tony Shen, Mohit Pandey, Jason Smith, Artem Cherkasov, and Martin Ester.
\newblock Tacogfn: Target conditioned gflownet for structure-based drug design.
\newblock \emph{arXiv e-prints}, pages arXiv--2310, 2023{\natexlab{a}}.

\bibitem[Ghari et~al.(2023)Ghari, Tseng, Eraslan, Lopez, Biancalani, Scalia, and Hajiramezanali]{ghari2023generative}
Pouya~M Ghari, Alex Tseng, G{\"o}kcen Eraslan, Romain Lopez, Tommaso Biancalani, Gabriele Scalia, and Ehsan Hajiramezanali.
\newblock Generative flow networks assisted biological sequence editing.
\newblock In \emph{NeurIPS 2023 Generative AI and Biology (GenBio) Workshop}, 2023.

\bibitem[Jain et~al.(2023{\natexlab{a}})Jain, Deleu, Hartford, Liu, Hernandez-Garcia, and Bengio]{jain2023gflownets}
Moksh Jain, Tristan Deleu, Jason Hartford, Cheng-Hao Liu, Alex Hernandez-Garcia, and Yoshua Bengio.
\newblock Gflownets for ai-driven scientific discovery.
\newblock \emph{Digital Discovery}, 2\penalty0 (3):\penalty0 557--577, 2023{\natexlab{a}}.

\bibitem[Hernandez-Garcia et~al.(2023)Hernandez-Garcia, Duval, Volokhova, Bengio, Sharma, Carrier, Koziarski, and Schmidt]{hernandezgarcia:hal-04407903}
Alex Hernandez-Garcia, Alexandre Duval, Alexandra Volokhova, Yoshua Bengio, Divya Sharma, Pierre~Luc Carrier, Micha{\l} Koziarski, and Victor Schmidt.
\newblock Crystal-gfn: sampling crystals with desirable properties and constraints.
\newblock In \emph{37th Conference on Neural Information Processing Systems (NeurIPS 2023)-AI4MAt workshop}, 2023.

\bibitem[Soares et~al.(2023)Soares, Cipcigan, Zubarev, and Brazil]{soares2023a}
Eduardo Soares, Flaviu Cipcigan, Dmitry Zubarev, and Emilio~Vital Brazil.
\newblock A framework for toxic pfas replacement based on gflownet and chemical foundation model.
\newblock In \emph{NeurIPS 2023 AI for Science Workshop}, 2023.

\bibitem[Volokhova et~al.(2023)Volokhova, Koziarski, Hern{\'a}ndez-Garc{\'\i}a, Liu, Miret, Lemos, Thiede, Yan, Aspuru-Guzik, and Bengio]{volokhova2023towards}
Alexandra Volokhova, Micha{\l} Koziarski, Alex Hern{\'a}ndez-Garc{\'\i}a, Cheng-Hao Liu, Santiago Miret, Pablo Lemos, Luca Thiede, Zichao Yan, Alan Aspuru-Guzik, and Yoshua Bengio.
\newblock Towards equilibrium molecular conformation generation with gflownets.
\newblock In \emph{AI for Accelerated Materials Design-NeurIPS 2023 Workshop}, 2023.

\bibitem[Jain et~al.(2023{\natexlab{b}})Jain, Raparthy, Hern{\'a}ndez-Garc{\i}a, Rector-Brooks, Bengio, Miret, and Bengio]{pmlr-v202-jain23a}
Moksh Jain, Sharath~Chandra Raparthy, Alex Hern{\'a}ndez-Garc{\i}a, Jarrid Rector-Brooks, Yoshua Bengio, Santiago Miret, and Emmanuel Bengio.
\newblock Multi-objective gflownets.
\newblock In \emph{International conference on machine learning}, pages 14631--14653. PMLR, 2023{\natexlab{b}}.

\bibitem[Deleu et~al.(2022)Deleu, G{\'o}is, Emezue, Rankawat, Lacoste-Julien, Bauer, and Bengio]{deleu2022bayesian}
Tristan Deleu, Ant{\'o}nio G{\'o}is, Chris Emezue, Mansi Rankawat, Simon Lacoste-Julien, Stefan Bauer, and Yoshua Bengio.
\newblock Bayesian structure learning with generative flow networks.
\newblock In \emph{Uncertainty in Artificial Intelligence}, pages 518--528. PMLR, 2022.

\bibitem[Zhang et~al.(2023{\natexlab{a}})Zhang, Rainone, Peschl, and Bondesan]{zhang2023robust}
David~W Zhang, Corrado Rainone, Markus Peschl, and Roberto Bondesan.
\newblock Robust scheduling with gflownets.
\newblock \emph{arXiv preprint arXiv:2302.05446}, 2023{\natexlab{a}}.

\bibitem[Zhang et~al.(2024)Zhang, Dai, Malkin, Courville, Bengio, and Pan]{zhang2024let}
Dinghuai Zhang, Hanjun Dai, Nikolay Malkin, Aaron~C Courville, Yoshua Bengio, and Ling Pan.
\newblock Let the flows tell: Solving graph combinatorial problems with gflownets.
\newblock \emph{Advances in Neural Information Processing Systems}, 36, 2024.

\bibitem[Kim et~al.(2023{\natexlab{a}})Kim, Yun, Bengio, Zhang, Bengio, Ahn, and Park]{kim2024local}
Minsu Kim, Taeyoung Yun, Emmanuel Bengio, Dinghuai Zhang, Yoshua Bengio, Sungsoo Ahn, and Jinkyoo Park.
\newblock Local search gflownets.
\newblock In \emph{The Twelfth International Conference on Learning Representations}, 2023{\natexlab{a}}.

\bibitem[Malkin et~al.(2022)Malkin, Jain, Bengio, Sun, and Bengio]{NEURIPS2022_27b51bac}
Nikolay Malkin, Moksh Jain, Emmanuel Bengio, Chen Sun, and Yoshua Bengio.
\newblock Trajectory balance: Improved credit assignment in gflownets.
\newblock \emph{Advances in Neural Information Processing Systems}, 35:\penalty0 5955--5967, 2022.

\bibitem[Ekbote et~al.(2022)Ekbote, Jain, Das, and Bengio]{ekbote2022consistent}
Chanakya Ekbote, Moksh Jain, Payel Das, and Yoshua Bengio.
\newblock Consistent training via energy-based gflownets for modeling discrete joint distributions, 2022.

\bibitem[Zhang et~al.(2023{\natexlab{b}})Zhang, Chen, Liu, Courville, and Bengio]{zhang2023diffusion}
Dinghuai Zhang, Ricky Tian~Qi Chen, Cheng-Hao Liu, Aaron Courville, and Yoshua Bengio.
\newblock Diffusion generative flow samplers: Improving learning signals through partial trajectory optimization.
\newblock \emph{arXiv preprint arXiv:2310.02679}, 2023{\natexlab{b}}.

\bibitem[Kim et~al.(2023{\natexlab{b}})Kim, Ko, Zhang, Pan, Yun, Kim, Park, and Bengio]{kim2023learning}
Minsu Kim, Joohwan Ko, Dinghuai Zhang, Ling Pan, Taeyoung Yun, Woochang Kim, Jinkyoo Park, and Yoshua Bengio.
\newblock Learning to scale logits for temperature-conditional gflownets.
\newblock \emph{arXiv preprint arXiv:2310.02823}, 2023{\natexlab{b}}.

\bibitem[Ma et~al.(2023)Ma, Bengio, Bengio, and Zhang]{ma2023baking}
George Ma, Emmanuel Bengio, Yoshua Bengio, and Dinghuai Zhang.
\newblock Baking symmetry into gflownets.
\newblock In \emph{NeurIPS 2023 AI for Science Workshop}, 2023.

\bibitem[Rector-Brooks et~al.(2023)Rector-Brooks, Madan, Jain, Korablyov, Liu, Chandar, Malkin, and Bengio]{rector2023thompson}
Jarrid Rector-Brooks, Kanika Madan, Moksh Jain, Maksym Korablyov, Cheng-Hao Liu, Sarath Chandar, Nikolay Malkin, and Yoshua Bengio.
\newblock Thompson sampling for improved exploration in gflownets.
\newblock \emph{arXiv preprint arXiv:2306.17693}, 2023.

\bibitem[Lahlou et~al.(2023)Lahlou, Deleu, Lemos, Zhang, Volokhova, Hern{\'a}ndez-Garc{\i}a, Ezzine, Bengio, and Malkin]{lahlou2023theory}
Salem Lahlou, Tristan Deleu, Pablo Lemos, Dinghuai Zhang, Alexandra Volokhova, Alex Hern{\'a}ndez-Garc{\i}a, L{\'e}na~N{\'e}hale Ezzine, Yoshua Bengio, and Nikolay Malkin.
\newblock A theory of continuous generative flow networks.
\newblock In \emph{International Conference on Machine Learning}, pages 18269--18300. PMLR, 2023.

\bibitem[Shen et~al.(2023{\natexlab{b}})Shen, Bengio, Hajiramezanali, Loukas, Cho, and Biancalani]{shen23accuracy}
Max~W Shen, Emmanuel Bengio, Ehsan Hajiramezanali, Andreas Loukas, Kyunghyun Cho, and Tommaso Biancalani.
\newblock Towards understanding and improving gflownet training.
\newblock In \emph{International Conference on Machine Learning}, pages 30956--30975. PMLR, 2023{\natexlab{b}}.

\bibitem[Madan et~al.(2023)Madan, Rector-Brooks, Korablyov, Bengio, Jain, Nica, Bosc, Bengio, and Malkin]{madan2023learning}
Kanika Madan, Jarrid Rector-Brooks, Maksym Korablyov, Emmanuel Bengio, Moksh Jain, Andrei~Cristian Nica, Tom Bosc, Yoshua Bengio, and Nikolay Malkin.
\newblock Learning gflownets from partial episodes for improved convergence and stability.
\newblock In \emph{International Conference on Machine Learning}, pages 23467--23483. PMLR, 2023.

\bibitem[Schulman et~al.(2017)Schulman, Wolski, Dhariwal, Radford, and Klimov]{schulman2017proximal}
John Schulman, Filip Wolski, Prafulla Dhariwal, Alec Radford, and Oleg Klimov.
\newblock Proximal policy optimization algorithms.
\newblock \emph{arXiv preprint arXiv:1707.06347}, 2017.

\bibitem[Xie et~al.(2020)Xie, Shi, Zhou, Yang, Zhang, Yu, and Li]{xie2021mars}
Yutong Xie, Chence Shi, Hao Zhou, Yuwei Yang, Weinan Zhang, Yong Yu, and Lei Li.
\newblock Mars: Markov molecular sampling for multi-objective drug discovery.
\newblock In \emph{International Conference on Learning Representations}, 2020.

\bibitem[Zhang et~al.(2020)Zhang, Liu, and Xie]{zhang2020molecular}
Shuo Zhang, Yang Liu, and Lei Xie.
\newblock Molecular mechanics-driven graph neural network with multiplex graph for molecular structures.
\newblock \emph{arXiv preprint arXiv:2011.07457}, 2020.

\bibitem[Sinai et~al.(2020)Sinai, Wang, Whatley, Slocum, Locane, and Kelsic]{sinai2021adalead}
Sam Sinai, Richard Wang, Alexander Whatley, Stewart Slocum, Elina Locane, and Eric~D Kelsic.
\newblock Adalead: A simple and robust adaptive greedy search algorithm for sequence design.
\newblock \emph{arXiv preprint arXiv:2010.02141}, 2020.

\bibitem[Trabucco et~al.(2022)Trabucco, Geng, Kumar, and Levine]{trabucco2022design}
Brandon Trabucco, Xinyang Geng, Aviral Kumar, and Sergey Levine.
\newblock Design-bench: Benchmarks for data-driven offline model-based optimization.
\newblock In \emph{International Conference on Machine Learning}, pages 21658--21676. PMLR, 2022.

\end{thebibliography}
\newpage
\appendix
\section{Details about GFNs}
The unnormalized density function $F$ can be metaphorically understood as the measure of water in the vast river system of directed acyclic Markov chains in GFN. It can represent the molarity, i.e., the number of water molecules, or the volume of water flow \cite{NEURIPS2022_27b51bac}. This analogy allows for comprehending the relationship between $F$ and the trajectory distribution probability $P$:
\begin{equation}\label{eq:Z}
        Z=F(s_0)=\sum_{\tau\in \mathcal{T}}F(\tau)
\end{equation}
From the definition of $Z$ in equation \ref{eq:Z}, it can be derived that each trajectory's probability is directly proportional to its unnormalized density function:
\begin{equation}
    P(\tau)=\frac{1}{Z}F(\tau)
\end{equation}
Next, one of the most important distinctions between GFN and conventional reinforcement learning \cite{bengio2021flow} is described, where the goal of GFN is to determine a Markov flow such that $P_F(x)$ is proportional to $R(x)$, as shown in equation \ref{eq:F eq R}.
\begin{equation}\label{eq:F eq R}
    F(s_0)=Z=\sum_{x \in \mathcal{X}} R(x)
\end{equation}
It should be noted that GFN distinguishes from classical reinforcement learning not only by prioritizing the reward value of the final compositional objects, which might be associated with a specific performance metric of a chemical compound, but also by emphasizing the approximation to the reward function distribution. This allows for the generation of compositional objects that may not score the highest on a particular metric but are more likely to be superior in other performance metrics. Balancing these two considerations is more valuable in practical production, and this is a key advantage of GFN over other RL approaches.The following equations presents the specific implementation objectives and corresponding loss functions of the two amortization algorithms used in the paper.

\textbf{Detailed balance:} From equations \ref{eq:PFtoF/F} and \ref{eq:PBtoF/F}, we have
\begin{equation}
     F(s)P_F(s_{t+1}\mid s_t)=F(s_{t+1})P_B(s_{t}\mid s_{t+1})
\end{equation}
Then, a loss function $\mathcal{L}_{DB}(s_t,s_{t+1})$ is designed to train the model parameters $\theta$ to achieve the goal of sampling where $P(x)$ is proportional to $R(x)$:
\[
\mathcal{L}_{DB}(s_t, s_{t+1}) = \left( \log \frac{F_\theta(s_t)P_F(s_{t+1}|s_t;\theta)}{F_\theta(s_{t+1})P_B(s_{t}|s_{t+1};\theta)} \right)^2
\]

\textbf{Trajectory balance:} Similarly, we can derive from \cite{NEURIPS2022_27b51bac}
\begin{equation}
Z \prod_{t=0}^{n} P_F(s_{t+1}|s_t) = F(x) \prod_{t=0}^{n} P_B(s_t|s_{t+1})
\end{equation}
Consequently, the following loss function $\mathcal{L}_{TB}$ is formulated:
\begin{equation}
\mathcal{L}_{TB}(\tau) = \left( \log \frac{Z_\theta \prod_{t=0}^{n-1} P_F(s_{t+1}|s_t;\theta)}{R(x) \prod_{t=0}^{n-1} P_B(s_t|s_{t+1};\theta)} \right)^2
\end{equation}
\section{Limitations and Future Work}
\paragraph{Limitations}
Although DB-GFN has demonstrated improvements in multiple performance aspects, it does not exhibit a great improvement on all GFN models and metrics. This includes the unique fraction shown in Figure \ref{fig:unique} and the fitting to the reward function. As demonstrated in Figure \ref{fig:old-ls-db acc} and Table \ref{tab:old-ls-db num}, although DB-GFN mostly considerably improves performance in already well-performing GFN networks, the gains in TB and MaxEnt are less pronounced for GFNs developed using the DB amortized inference algorithm and subTB. This is mainly because, even though DB-GFN can enhance model performance through reward-based dynamic backtracking, it relies on the original model's exploratory ability in the sampling space. If the exploratory capability is weak, even after backtracking, DB-GFN is still not easy to find good solutions.Moreover, for the generative model DB-GFN, there are risks of negative \textbf{broader impacts}, such as the potential misuse in developing highly toxic poisons or addictive substances. These risks can be mitigated through restricted access and stringent control over related raw materials and datasets.
\paragraph{Future Work}
For future work, there is potential to discover new selection algorithms that may yield better results. Although the experiments in this paper primarily set the probability of backtracking comparison in DB-GFN at 100\%, introducing randomness could still enhance sampling outcomes for specific tasks.
\section{Detailed Experimental Setting}
\subsection{Hyper-parameters}
\begin{table}
    \centering
    \begin{adjustbox}{width=0.6\textwidth}
    \begin{tabular}{ccccccc}
        \toprule
         hyperparameters&  QM9 &  sEH &  TFbind8 &  RNA1 &  RNA2 &  RNA3\\
         \midrule
         $S_m$ & $5$ &  $5$ & $6$ & $9$ & $9$ & $9$ \\
         $S_l$ & $2$ &  $2$ & $2$ & $5$ & $5$ & $5$ \\
         $T_m$ & $5$ &  $5$ & $5$ & $5$ & $5$ & $5$ \\
         $T_l$ & $0.4$ &  $0.4$ & $0.4$ & $0.4$ & $0.4$ & $0.4$ \\
         \bottomrule
    \end{tabular}
    \end{adjustbox}
    \caption{Key Hyperparameters of DB-GFN}
    \label{tab:hyper}
\end{table}
In the multiple comparative experiments conducted in this study, to ensure fairness and adequately show the performance of previous works, we strictly adhered to the hyperparameters and training processes previously published. Additionally, the development of DB-GFN employed the same training methodologies as earlier studies, including the use of prioritized replay training and relative edge flow policy parametrization mapping.

The key hyperparameters of DB-GFN are outlined in Table \ref{tab:hyper}. For comprehensive explanations of parameters such as learning rate and batch size, please consult the project's script files and configuration documents. For access to the anonymous version of the project, visit: \url{https://anonymous.4open.science/r/db_gfn}.
\subsection{Additional Experiment}
\begin{figure}
    \centering
    \includegraphics[width=0.95\linewidth]{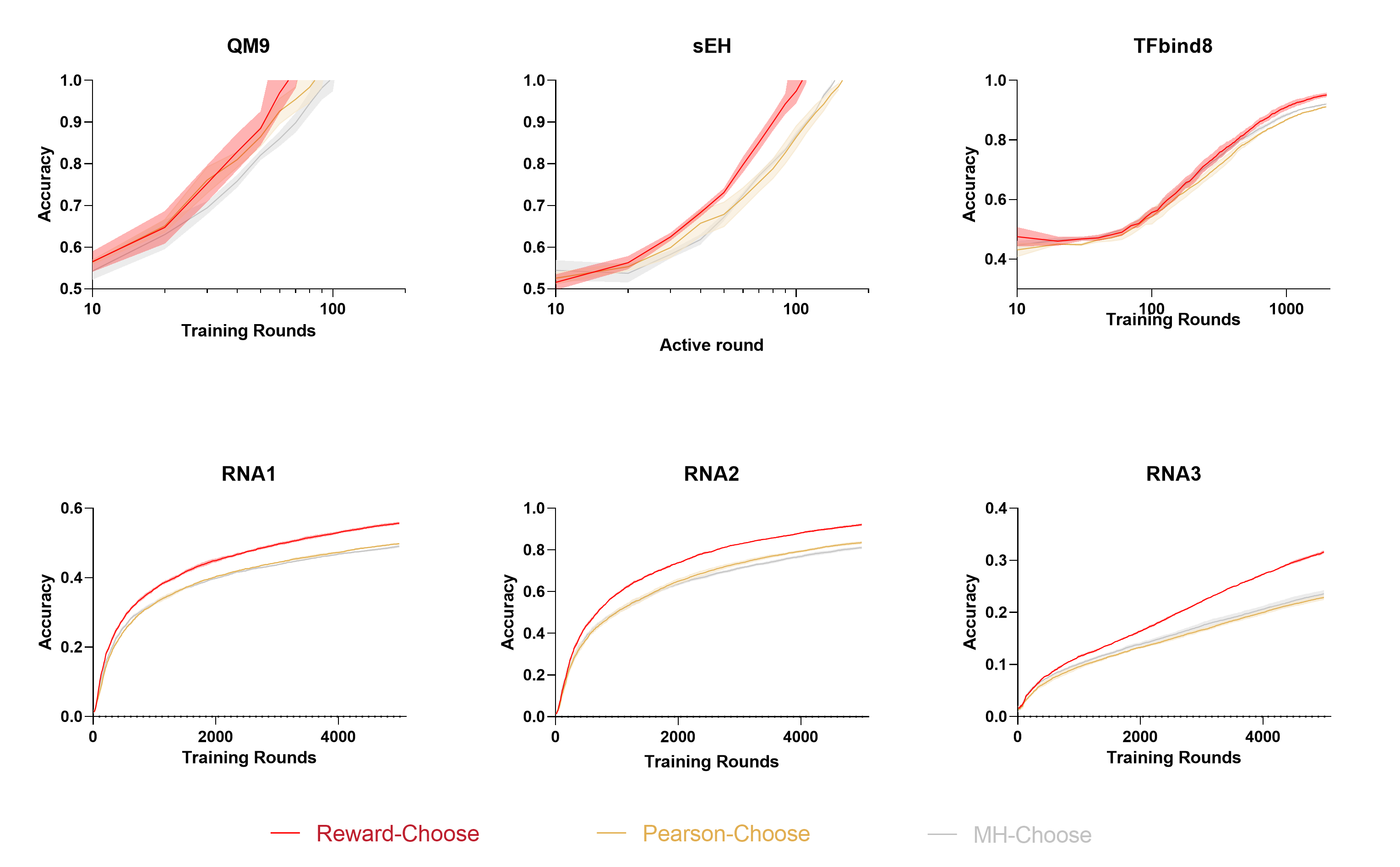}
    \caption{Accuracy of three selection algorithms}
    \label{fig:choose_acc}
\end{figure}
Generative models primarily focus on the performance of generated samples, specifically the magnitude of the final reward value. It can be anticipated that using the Reward-Choose selection algorithm would significantly enhance the performance improvements facilitated by DB-GFN+TB, as detailed in Figure \ref{fig:choose_acc}. It is evident that all three selection algorithms improve performance, with the Reward-Choose algorithm showing particularly notable enhancements.
\end{document}